# Detecting Blackholes and Volcanoes in Directed Networks

Zhongmou Li, Hui Xiong, Yanchi Liu
Management Science and Information Systems Department
Rutgers, the State University of New Jersey
E-mail: mosesli@pegasus.rutgers.edu, hxiong@rutgers.edu, yanchi@rutgers.edu

## ABSTRACT

In this paper, we formulate a novel problem for finding blackhole and volcano patterns in a large directed graph. Specifically, a blackhole pattern is a group which is made of a set of nodes in a way such that there are only inlinks to this group from the rest nodes in the graph. In contrast, a volcano pattern is a group which only has outlinks to the rest nodes in the graph. Both patterns can be observed in real world. For instance, in a trading network, a blackhole pattern may represent a group of traders who are manipulating the market. In the paper, we first prove that the blackhole mining problem is a dual problem of finding volcanoes. Therefore, we focus on finding the blackhole patterns. Along this line, we design two pruning schemes to guide the blackhole finding process. In the first pruning scheme, we strategically prune the search space based on a set of pattern-size-independent pruning rules and develop an iBlackhole algorithm. The second pruning scheme follows a divide-and-conquer strategy to further exploit the pruning results from the first pruning scheme. Indeed, a target directed graphs can be divided into several disconnected subgraphs by the first pruning scheme, and thus the blackhole finding can be conducted in each disconnected subgraph rather than in a large graph. Based on these two pruning schemes, we also develop an iBlackhole-DC algorithm. Finally, experimental results on real-world data show that the iBlackhole-DC algorithm can be several orders of magnitude faster than the iBlackhole algorithm, which has a huge computational advantage over a brute-force method.

## 1. INTRODUCTION

Financial institutions and government agencies, such as U.S. Securities and Exchange Commission (SEC), are facing some daunting challenges in the field of financial fraud detection. The sophistication of criminals' tactics makes detecting and preventing fraud difficult, especially as the number of trading accounts and the volume of transactions grow dramatically. Indeed, the trading networks are vulnerable to these fast-growing accounts and the volume of transactions. Particularly, criminals know fraud detection systems are not good at correlating user behavior across multiple trading accounts. This weakness opens the door for cross-account collaborative fraud, which is difficult to discover, track and resolve because the activities of the fraudsters usually appear to be normal trading activities. For instance, consider a trading network with a large number of nodes and directed edges, a trader or a group of traders can perform trading only within several accounts for the purpose of manipulating the market. This kind of illegal trading activities is widely known as trading ring.

In this paper, we study a special type of trading-ring patterns, called blackhole and volcano patterns. Given a directed graph, a blackhole pattern is a group which is made of a set of nodes in a way such that there are only inlinks to this group from the rest nodes in the graph. In contrast, a volcano pattern is a group which only has outlinks to the rest nodes in the graph. To the best of our knowledge, this is the first time to have the concepts of blackhole and volcano patterns in the directed graphs. In fact, both blackhole and volcano patterns can be observed in real-world trading networks. For example, a blackhole pattern can represent a group of traders who are manipulating the market by performing transactions on a specific stock among themselves for a specific time period. In other words, the overall shares of the target stock in their trading accounts can only increase during this time period, while these traders have produced a large volume of transactions on this stock. After the stock price goes up to a certain degree, these traders start selling off their shares to the public. In this stage, these trading accounts form a volcano pattern which only has outlines to the rest public accounts.

However, the process for finding blackhole and volcano patterns can be computationally prohibited, since this is a combinatorial problem in nature. To address this challenge, we first prove that the blackhole pattern mining problem is a dual problem of finding volcano patterns. Therefore, we can focus on finding the blackhole patterns. Along this line, we design two pruning schemes to guide the blackhole pattern mining process. In the first pruning scheme, we identify a set of pattern-size-independent pruning rules by studying the structural graph properties of blackhole patterns. These pruning rules can be used for pruning the search space no matter the size of the patterns is. Based on the first pruning scheme, we design an iBlackhole algorithm for finding blackhole patterns. In contrast, the second pruning scheme follows a divide-and-conquer strategy to further exploit the pruning results from the first pruning scheme. Specifically, because a target directed graph have been divided into several disconnected subgraphs by the first pruning scheme, it becomes much more efficient to find blackhole patterns in each disconnected subgraphs rather than in a large graph. Based on these two pruning schemes, we develop an even more effective algorithm, named iBlackhole-DC, for mining blackhole patterns in directed graphs. Furthermore, we have provided the proof of the completeness and correctness of both iBlackhole and iBlackhole-DC algorithms.

Finally, experiments results on several real-world directed networks are provided to show the pruning effect of two pruning schemes. As shown in the experiments, the iBlackhole algorithm has a huge computational advantage over a brute-force approach. Also, the iBlackhole-DC algorithm is several orders of magnitude faster than the iBlackhole algorithm. Finally, we show the effectiveness of blackhole patterns for finding some interesting stock movement patterns.

## 2. PRELIMINARIES

In this section, we introduce some basic concepts and notations that will be used in this paper.

First, consider a directed graph $G = (V, E)$ [5], where $V$ is the set of all nodes and $E$ is the set of all edges. Assume that $G$ has no self-loop and has no more than one edge between any pair of nodes. A directed edge $e$ in $G$ is denoted as $e = (x, y)$, where $x$ and $y$ are nodes of $G$ and an arc is directed from $x$ to $y$. Each edge $e$ has a positive weight, denoted as $\omega_e$, associated with this edge.

**Definition 1 (*in-weight and out-weight*).** For a set of nodes $B \subseteq V$, and let $C = V \setminus B$, the *in-weight* of $B$ is defined as: $d_{in}(B) = \sum_{e=(x,y)\in E, x\in C, y\in B} \omega_e$ . And the definition of the *out-weight* of $B$ is very similar: $d_{out}(B) = \sum_{e=(x,y)\in E, x\in B, y\in C} \omega_e$ .

Figure 1 shows an example of the in-weight and out-weight of a set of nodes. The number associated with each edge is the weight of that edge. In this figure, the in-weight of $B$ is 6 + 5 = 11, while the out-weight is 3 + 3 + 1 + 2 = 9.

Next, we give the definition of the blackhole in a directed graph as the following.

**Definition 2 (*blackhole*).** Given a directed graph $G = (V, E)$, we say that a set of nodes $B \subseteq V$ form a *blackhole*, if and only if the following two conditions are satisfied: 1) If $|B| \geqslant 2$, the subgraph $G(B)$ induced by $B$ is weakly connected, and 2) $d_{in}(B) / d_{out}(B) > \theta$, where $\theta$ is a pre-specified positive threshold and is typically a very large value.

Finally, we present the definition of the volcano in a directed graph as follows.

**Definition 3 (*volcano*).** Given a directed graph $G = (V, E)$, we say that a set of nodes $Vol \subseteq V$ form a *volcano,* if and only if the following two conditions are satisfied: 1) If $|B| \geqslant 2$, the subgraph $G(Vol)$ induced by $Vol$ is weakly connected, and 2) $d_{out}(Vol) / d_{in}(Vol) > \theta$, where $\theta$ is a pre-specified positive threshold and is typically a very large value.

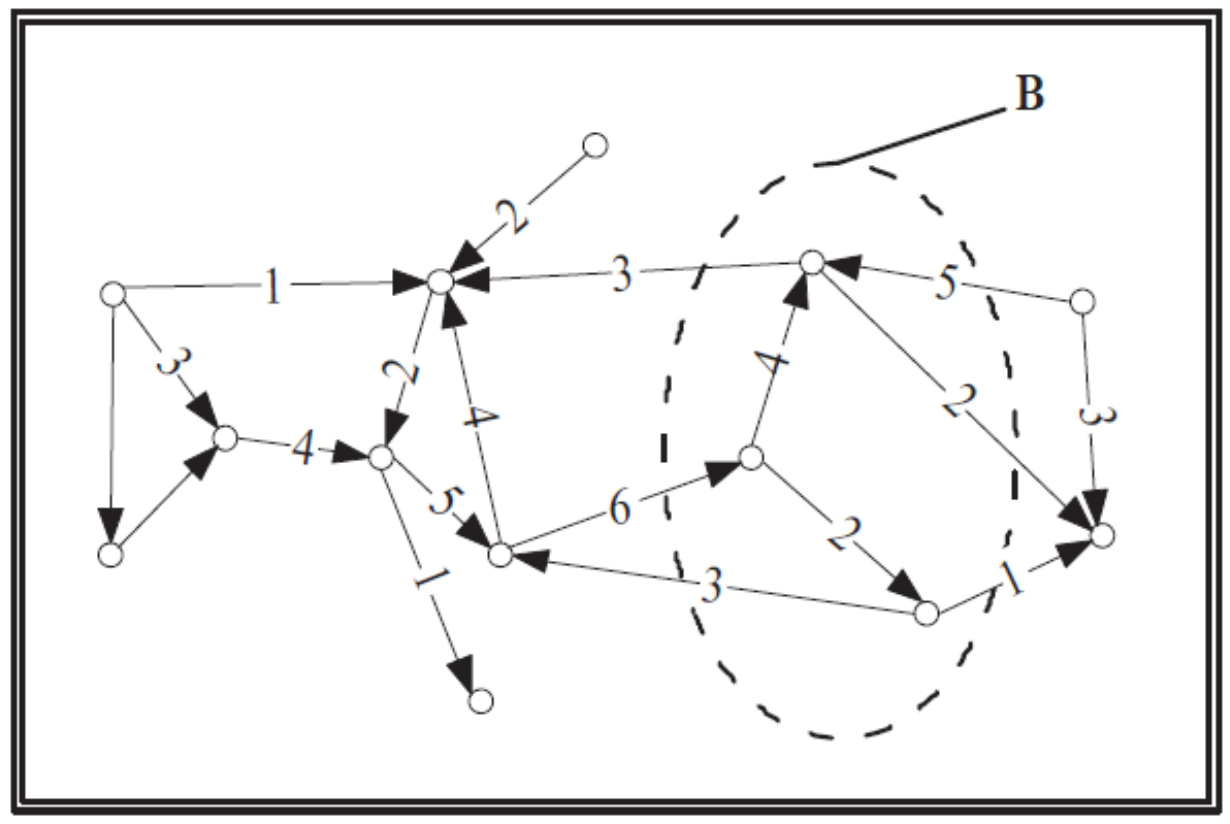

**Figure 1: Illustration: in-weight and out-weight**

## 3. PROBLEM FORMULATION

In this section, we formulate the problems of detecting blackhole and volcano patterns in a directed graph.

### 3.1 A General Problem Formulation

Given a directed graph $G = (V, E)$, the goal of detecting blackhole patterns is to find out the blackhole set, denoted as *Blackhole*, such that, 1) for each element $B \in Blackhole$, $B \subseteq V$ and $B$ satisfies the definition of *blackhole*, and 2) for any other set of nodes $C \subseteq V$ and $C \notin Blackhole$, $C$ does not satisfy the *blackhole* definition. The problem of detecting volcano patterns can be formulated in a similar fashion.

Next, we show that the problem of detecting blackhole patterns is a dual problem of detecting volcano patterns.

**Theorem 1.** The problem of finding out the blackhole set in a directed graph is a dual problem of finding out the volcano set in the same directed graph.

**Proof.** Consider a directed graph $G = (V, E)$. Let $G' = (V, E')$ be the inverse graph of $G$, where all the nodes in $G'$ are the same as in $G$; while for each edge $e = (x, y) \in E$, there is an edge $e' = (y, x) \in E'$, and the weight associated with $e'$ are exactly the same as the weight associated with $e$. Therefore, the in-weight of a set of nodes $B$ in $G$ are exactly the same as the out-weight of $B$ in $G'$, and vice versa. If $B$ is a blackhole in $G$, which means $d_{in}(B) / d_{out}(B) > \theta$ in $G$, then in $G'$, we have $d_{out}(B) / d_{in}(B) > \theta$. Therefore, $B$ forms a volcano in $G'$. As a result, the problem of finding out the blackhole set in $G$ is equivalent to the problem of finding out the volcano set in $G'$. □

Now that we know the problem of detecting the blackhole set in the original directed graph is equivalent to the problem of detecting the volcano set in the inverse graph. Therefore, in the rest of this paper, we can only focus on detecting blackhole patterns in a directed graph.

### 3.2 A Simplified Problem Formulation

The above general problem of detecting blackhole patterns is very complex. Instead, in this paper, we focus on a more practical version of this problem. Specifically, we exploit two constraints to simplify the general problem as follows. 1) The weights associated with all edges are all equal to 1. This constraint results that the in-weight of a node becomes the in-degree and the out-weight becomes the out-degree; 2) Instead of considering the general version of a blackhole, which satisfies $d_{in}(B) / d_{out}(B) > \theta$, we simplify the blackhole definition with $d_{out}(B) = 0$.

Figure 2 shows an example of the simplified blackhole patterns. In this figure, there are two blackhole patterns, which have been highlighted by dashed circles.

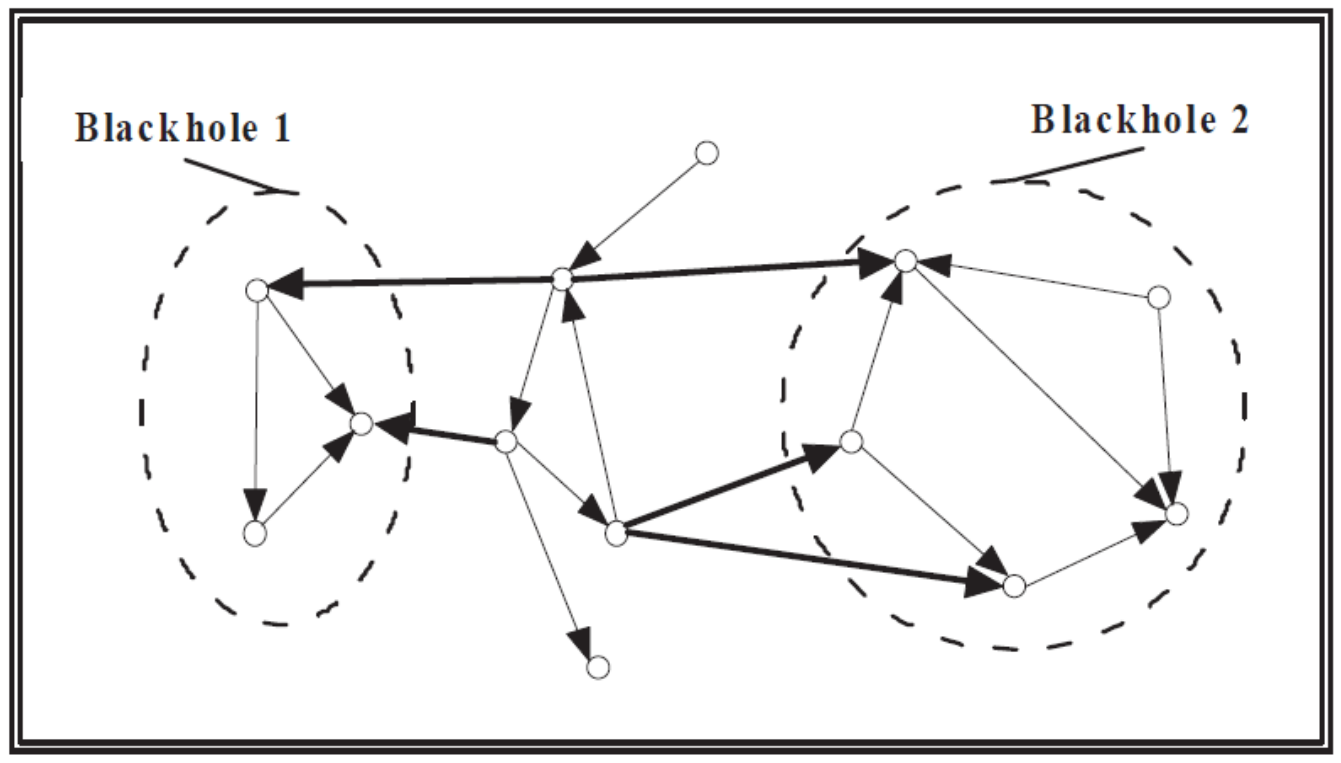

**Figure 2: An illustration of the simplified blackhole**

## 4. ALGORITHM DESIGN

In this section, we introduce the algorithms for detecting blackhole patterns in a directed graph.

### 4.1 A Brute-Force Approach

First, we present a brute-force approach for finding blackhole patterns. As we know, a set of nodes $B \subseteq V$ is a blackhole, if and only if: 1) If $|B| \geqslant 2$, the subgraph $G(B)$ induced by $B$ is weakly connected, and 2) $d_{out}(B) = 0$. Therefore, the intuition is really simple: all the possible combinations of the nodes in $G$ are checked using the exhaustive search method. Also, for each combination $B$, if the subgraph $G(B)$ induced by $B$ is weakly connected and $d_{out}(B) = 0$, then $B$ is a blackhole in the directed graph $G$.

In real-world scenarios, it is typically computational prohibited to find all the blackhole patterns, since the number of combinations of the nodes is exponentially increased as the number of nodes. A practical way is to find blackhole patterns which include only limited number of nodes. Here, we introduce a concept of *n-node blackholes*.

**Definition 4 (*n-node blackhole*).** Given a directed graph $G = (V, E)$, we say that a set of nodes $B \subseteq V$ is an *n-node blackhole*, if and only if the following two conditions are satisfied: 1) $B$ is a blackhole in $G$, and 2) $B \in V(n)$, where $V(n)$ is the set of all possible subsets containing n nodes in $V$; that is, $|B| = n$.

Figure 3 shows the pseudocode of the brute-force algorithm to detect 1 through $n$-node blackhole patterns in a directed graph $G$. Since we have considered all the possible combinations of nodes in $G$ from 1 through n, this algorithm is complete. Also, since for each combination of nodes, we have checked whether it satisfies the definition of blackhole, this algorithm is correct.

```
ALGORITHM BRUTE-FORCE(G = (V, E), n)
Input:
        G: the directed graph
        V: the set of all nodes
        E: the set of all edges
        n: max number of nodes each blackhole may contain
Output:
        Blackhole: 1 to n-node blackhole set of G

        Blackhole ← ∅
        for i ← 1 to n do
            for each B ∈ V(i) do
                if G(B) is weakly connected then
                    if d_out(B) == 0 then
                        Blackhole ← Blackhole ⋃ B
                    end if
                end if
            end for
        end for
        return Blackhole
```

**Figure 3: The brute-force algorithm**

## 4.2 A Scheme of the iBlackhole Algorithm

In general, finding blackhole patterns in a directed graph is a combinatorial problem. Therefore, as the number of nodes $n$ increases, the computation time increases exponentially, making the brute-force algorithm unrealistic to obtain the result for a large $n$ value. To this end, we introduce some pattern-size-independent pruning rules to reduce the search space. The key idea behind these pruning rules is to find out irrelevant nodes that have no chance to form an $n$-node blackhole as many as possible, and eliminate these nodes from the candidate search list. In this way, the search space can be reduced dramatically. The algorithm developed based on these pruning rules is named as *iBlackhole.*

Figure 4 shows the scheme of the iBlackhole algorithm for detecting 1 through $n$-node blackhole patterns in a directed graph $G$. In this algorithm, all the blackhole patterns are identified one by one according to their number of nodes. In each step of finding the $i$-node blackhole patterns, a potential list $P_i$ is first established. Only the nodes in this potential list have possibilities to form an $i$-node blackhole. In other words, nodes that are not in this list have no chance to be in an $i$-node blackhole pattern. Then nodes in $P_i$ will be examined one after another and irrelevant nodes will be deleted based on some pruning rules. The results of this pruning form a candidate list $C_i$. For each node $v$ in $C_i$, we will check it again and remove irrelevant nodes from $C_i$ using some additional pruning rules. Finally, we will have the final search list $F_i$, and then we can apply the brute-force algorithm on $F_i$ to find out the $i$-node blackhole patterns. More details about this algorithm will be given after we introduce some pruning rules.

## 4.3 Pruning Rules

In this section, we introduce pruning rules associated with the potential list, the candidate list, and the final search list.

**Definition 5 (*directed path*).** Given a directed graph $G = (V, E)$, $v_0, v_1, v_2, \dots, v_k \in V$, $e_1, e_2, \dots, e_k \in E$, where $e_i = (v_{i-1}, v_i)$. We say that the sequence of $v_0e_1v_1e_2v_2 \dots e_kv_k$ forms a *directed path* from $v_0$ to $v_k$, if $v_i \neq v_j$ for all $0 \leqslant i, j \leqslant k$, $i \neq j$. The length of this directed path is $k$.

**Definition 6 (*reachable*).** Given a directed graph $G = (V, E)$, $u, v \in V$. We say that $v$ is *reachable* from $u$ if there is a directed path that starts from $u$ and ends at $v$.

**Definition 7 (*predecessor and successor*).** Given a directed graph $G = (V, E)$, $u, v \in V$. If $v$ is reachable from $u$, then we say $u$ is a *predecessor* of $v$, and $v$ is a *successor* of $u$. If there is an edge from $u$ to $v$, then $u$ is a *direct predecessor* of $v$, and $v$ is a *direct successor* of $u$.

```
ALGORITHM iBlackhole(G = (V, E), n)
Input:
        G: the directed graph
        V: the set of all nodes
        E: the set of all edges
        n: max number of nodes each blackhole may contain
Output:
        Blackhole: 1 to n-node blackhole set of G

        Blackhole ← ∅
        for i ← 1 to n do
            establish potential list P_i
            remove irrelevant nodes from P_i, get candidate list C_i
            remove irrelevant nodes from C_i, get final list F_i
            apply the Brute-Force Algorithm on F_i
              to find out the i-node blackhole patterns
        end for
        return Blackhole
```

**Figure 4: A scheme of the iBlackhole algorithm**

**Lemma 1.** If a node $v \in B$, where $B \subseteq V$ is a blackhole, then all the direct successors of $v$ are all in $B$.

**Proof.** This can be proved by contradiction. Assume that there is at least one of $v$'s direct successors $s$, and $s \notin B$, then we have $d_{out}(B) \geqslant 1$ *since* $e = (v, s) \in E$ and $s \notin B$. This contradicts with the definition of blackhole. Therefore, all direct successors of $v$ should be in B. □

Based on Lemma 1, we have the following lemma.

**Lemma 2.** In an $n$-node blackhole $B$, the maximum out-degree of any node in $B$ is $n-1$.

**Proof.** This can be proved by contradiction. Suppose there is a node $v$ with out-degree at least $n$ in an $n$-node blackhole $B$, then $v$ should have at least $n$ direct successors, denoted as $s_1, s_2, \ldots, s_n$. According to Lemma 1, if $v \in B$, all of $s_1, s_2, \ldots, s_n$ should be in $B$, which makes the size of this blackhole at least $n+1$. Then we find a contradiction here. Therefore, the maximum out-degree of any node in an $n$-node blackhole should be no greater than $n-1$. □

According to Lemma 2, we can derive the following theorem for pruning the potential list $P_i$.

**Theorem 2.** For the potential list $P_i$, only nodes with out-degree less than $i$ need to be considered.

**Proof.** By Lemma 2, in an $n$-node blackhole, the maximum out-degree of any node is $n-1$. In other words, nodes with out-degree greater than $i-1$ have no chance to be in an $i$-node blackhole. Therefore, only nodes with out-degree less than $i$ needs to be included in the potential list $P_i$. □

According to Theorem 2, only the nodes with out-degree less than $i$ are used to establish the potential list $P_i$. After having $P_i$, some additional pruning rules can be applied to remove irrelevant nodes from $P_i$ to get the candidate list $C_i$.

**Lemma 3.** For each node $v \in P_i$, if there is at least one of $v$'s direct successors $s \notin P_i$, then $v \notin C_i$.

**Proof.** This can be proved by contradiction. Since $s \notin P_i$, this means s has no chance to be in an $i$-node blackhole. Assume that finally $v$ belongs to an $i$-node blackhole $B$. According to Lemma 1, all $v$'s direct successors, which include $s$, will also belong to $B$. Then we have a contradiction here. Therefore, $v$ has no chance to form an $i$-node blackhole, and thus $v$ can be removed from $P_i$ safely; that is, $v \notin C_i$. □

After a node is removed from $P_i$, there are some other nodes associated with it can also be removed from $P_i$.

**Lemma 4.** If a node $v$ is removed from $P_i$, then all of its direct predecessors can also be removed from $P_i$.

**Proof.** For each of $v$'s direct predecessors $p$, $v$ is $p$'s direct successor. Since $v$ has been removed from $P_i$, then $v \notin P_i$. According to Lemma 3, $p$ should also be removed from $P_i$. Therefore, all $v$'s direct predecessors can be removed from $P_i$. □

By Lemma 4, when removing a node $v$ from $P_i$, all its direct predecessors should also be removed. Then, the newly removed direct predecessors become the new *"v"*s. Finally, the cascading delete will spread to all $v$'s predecessors. Figure 5 shows an example of the cascading delete process when removing node $v$ from the potential list $P_3$. The shadow nodes in the figures are nodes removed from $P_3$. In Figure (a), $s$ has an out-degree of 3, which makes it exclude from $P_3$ at the first place. When nodes in $P_3$ are checked one after another, it can be noticed that $v$ has a successor $s$ not in $P_3$. Therefore, $v$ is removed from $P_3$. Then all of $v$'s direct predecessors are all deleted from $P_3$ as shown in Figure (b), and this process spreads to all $v$'s predecessors in Figure (c).

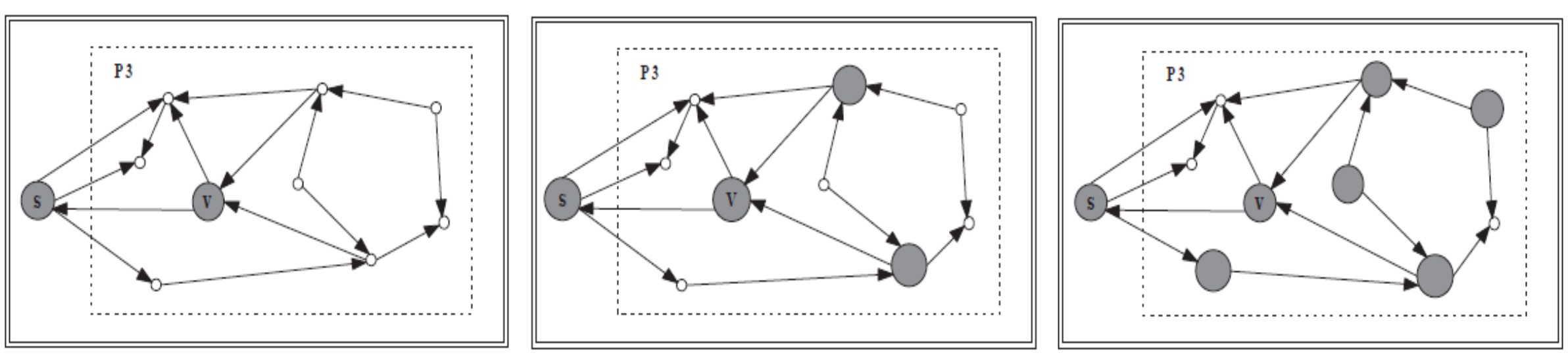

(a) (b) (c)

**Figure 5: Illustration: the cascading delete process**

Lemma 3 and Lemma 4 are the pruning rules which are used on $P_i$ to get the candidate list $C_i$. Nonetheless, some of the nodes in $P_i$ do not need to be examined and will definitely be in $C_i$.

**Lemma 5.** If a node $v \in C_{i-1}$, then $v \in C_i$.

**Proof.** Clearly, $C_{i-1} \subseteq P_{i-1}$, and $P_{i-1} \subseteq P_i$. Therefore, $C_{i-1} \subseteq P_i$. In other words, if $v \in C_{i-1}$, $v \in P_i$. Since all the nodes in $C_{i-1}$ only point to other nodes in $C_{i-1}$, all their successors are still in $C_{i-1}$. Therefore, when pruning rules (Lemma 3 and Lemma 4) are applied to $P_i$, there is no chance for $v$ to be removed from $P_i$ by these pruning rules. Finally, we know $v \in C_i$. □

According to Lemma 5, there is no need to examine nodes in $C_{i-1}$ when applying pruning rules to $P_i$, which makes this step more efficient.

Before we can continue to introduce the pruning strategies, we would like to introduce another concept here.

**Definition 8 (closure).** Given a directed graph $G = (V, E)$, $v \in V$. The *closure* of $v$, denoted as $v^+$, is defined as: $v^+ = \{s \mid \text{there is a directed path from } v \text{ to } s\} \cup \{v\}$.

Figure 6 shows an example of the closure of node $v$. In this figure, $v^+ = \{v, a, c, b, h, g\}$.

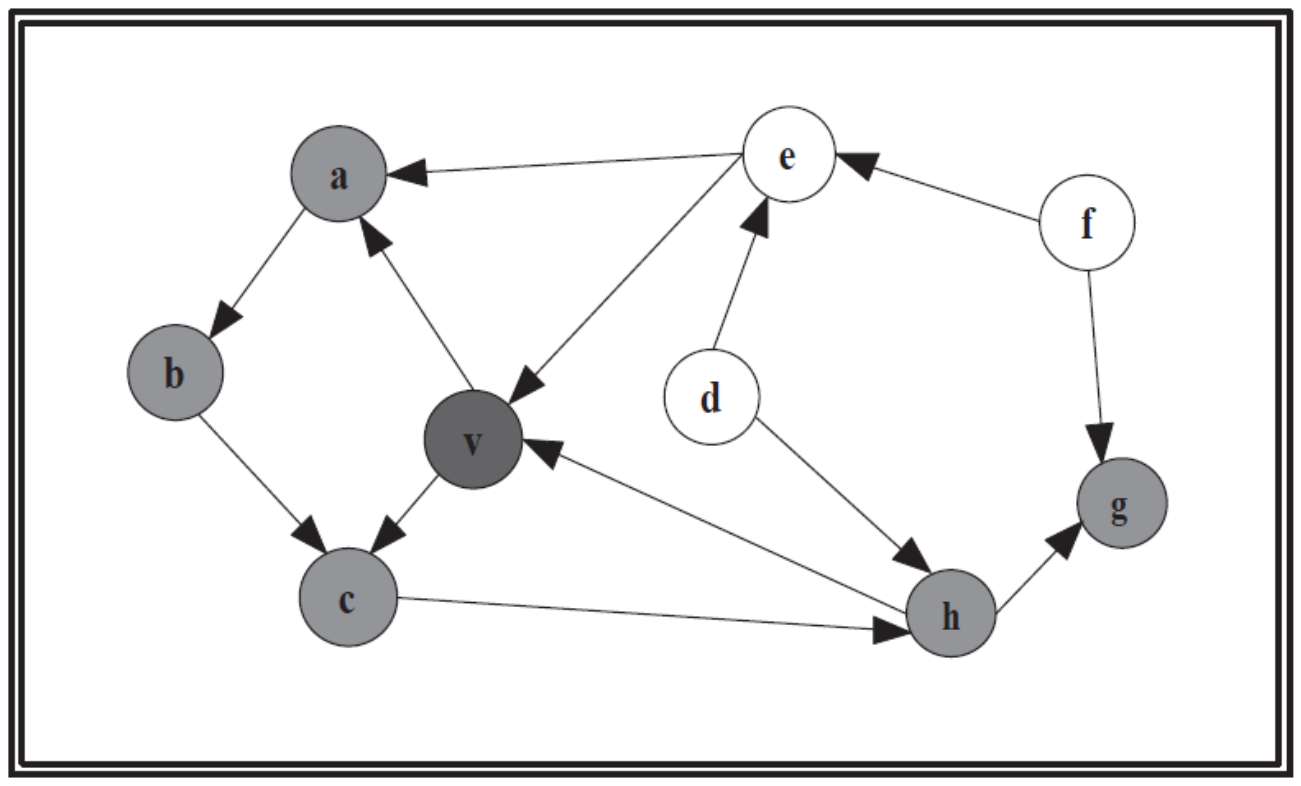

**Figure 6: An example of the closure of node *v***

Indeed, the closure of a node has an important feature as the following.

**Theorem 3.** The closure of a node $v$ is a blackhole. Furthermore, it is a subset of any blackhole which contains $v$.

**Proof.** For the first part of this theorem, by the definition of closure, $v^+$ is the set of all nodes reachable from $v$, together with $v$. If $v^+$ does not form a blackhole, there have to be at least an edge $e = (s, t) \in G$, such that $s \in v^+$ and $t \notin v^+$. If $s = v$, then $t$ is reachable from $v$, we have $t \in v^+$; If $s \neq v$, since there is a directed path from $v$ to $s$, and $e = (s, t)$, $t$ can be reached from $v$. We can also have $t \in v^+$. In either condition, we can have a contradiction here. Therefore, $v^+$ is a blackhole.

For the second part of this theorem, if a blackhole $B$ contains $v$, by Lemma 1, all $v$'s direct successors should all be in $B$. And then these direct successors become the new *"v"*s. Eventually, this procedure will be spread to all the $v$'s successors. The above leads to $v^+ \subseteq B$. □

The feature of closure (Theorem 3) can be used to derive some pruning rules to remove some irrelevant nodes from $C_i$, and finally lead to the final search list $F_i$.

**Lemma 6.** For each node $v \in C_i$, if $|v^+| > i$, then $v$ and all its predecessors are not in $F_i$.

**Proof.** By Theorem 3, $v^+$ is a subset of any blackhole which contains $v$. Suppose $v$ is in an $i$-node blackhole $B$. Then we have $v^+ \subseteq B$. So $|B| \geqslant |v^+| > i$. We can have a contradiction here. Therefore, $v$ has no chance to form an $i$-node blackhole, and we can remove $v$ from $C_i$ safely. Then, the similar cascading delete procedure can be applied, and thus all the $v$'s predecessors can be deleted from $C_i$. Therefore, $v$ and all its predecessors will not be in $F_i$. □

**Lemma 7.** For each node $v \in C_i$, if $|v^+| = i$, then $v^+$ can be outputted as an $i$-node blackhole. Also, $v$ and all its predecessors can be removed from $C_i$.

**Proof.** According to Theorem 3, $v^+$ is a blackhole. Since $|v^+| = i$, $v^+$ can be outputted as an $i$-node blackhole. Assume that $v$ will also be in another blackhole $B$. By Theorem 3, $v^+ \subseteq B$. If $|B| > i$, $B$ is not an $i$-node blackhole and cannot be outputted as an $i$-node blackhole; If $|B| = i$, then $B$ is exactly $v^+$, and has already been out putted as an $i$-node blackhole. In either situation, we can remove $v$ from $C_i$. Then the similar cascading delete procedure can be applied, and thus all $v$'s predecessors can be deleted from $C_i$. □

Lemma 6 and Lemma 7 are used as pruning rules to prune the candidate list $C_i$ and get the final search list $F_i$. After having $F_i$, we can apply the brute-force approach on $F_i$ to find out the $i$-node blackhole patterns. In the next subsection, we will give the details of the iBlackhole algorithm.

### 4.4 The iBlackhole Algorithm

The iBlackhole algorithm exploits the pruning rules stated from Lemma 1 to Lemma 7. Figure 7 shows the detailed pseudocode of the iBlackhole algorithm. Specifically, Line 4 establishes the potential list $P_i$. Lines 5 - 14 remove irrelevant nodes from $P_i$, and get the candidate list $C_i$. Lines 15 - 26 remove irrelevant nodes from $C_i$, and get the final search list $F_i$. Lines 27 - 33 apply the brute-force approach on $F_i$ to find out the $i$-node blackhole patterns.

```
ALGORITHM iBlackhole(G = (V, E), n)
Input:
        G: the directed graph
        V: the set of all nodes
        E: the set of all edges
        n: max number of nodes each blackhole may contain
Output:
        Blackhole: 1 to n-node blackhole set of G

 1.     Blackhole ← ∅
 2.     C_0 ← ∅
 3.     for i ← 1 to n do
 4.         P_i ← {v|d_out(v) < i}
 5.         for each v in P_i do
 6.             if v ∉ C_{i−1} then
 7.                 if at least one of v's directed
 8.                   successors are not in P_i then
 9.                     remove v from P_i
10.                     remove all v's predecessors from P_i
11.                 end if
12.             end if
13.         end for
14.         C_i ← P_i
15.         for each v in C_i do
16.             if |v+| > i then
17.                 remove v from C_i
18.                 remove all v's predecessors from C_i
19.             end if
20.             if |v+| == i then
21.                 Blackhole ← Blackhole ⋃ v+
22.                 remove v from C_i
23.                 remove all v's predecessors from C_i
24.             end if
25.         end for
26.         F_i ← C_i
27.         for each B ∈ F_i(i) do
28.             if G(B) is weakly connected then
29.                 if d_out(B) == 0 then
30.                     Blackhole ← Blackhole ⋃ B
31.                 end if
32.             end if
33.         end for
34.     end for
35.     return Blackhole
```

**Figure 7: The iBlackhole algorithm**

**Completeness and Correctness.** In the iBlackhole algorithm, since only the nodes that have no chance to form an *i*-node blackhole pattern are removed in each iteration *i* (this is guaranteed by Lemma 1 through Lemma 7). In other words, all the possible combinations of nodes have been checked to produce $F_i$, this algorithm is complete. Also, for each candidate blackhole pattern, since we have checked whether this candidate pattern satisfies the definition of blackhole or not, this algorithm is correct.

Figure 8 shows an example of the procedure of the iBlackhole algorithm when searching the 3-node blackhole patterns. The shadow nodes in the figures are nodes which have been deleted. In Figure (a), *s* has an out-degree of 3, so *s* can be deleted from $P_3$ at the first place. In Figure (b), when we check the nodes in $P_3$ one after another, we notice that *v* has a successor *s* not in $P_3$. Therefore, we can remove *v* from $P_3$. Then, all the direct predecessors of *v* can be cascaded deleted from $P_3$, and this delete process spreads to all *v*'s predecessors. Finally, we have the candidate list $C_3 = \{a, b, c, i, j, k\}$. In Figure (c), we find that $|i^+| = 3$. Therefore, we output $i^+ = \{i, j, k\}$ as a 3-node blackhole, and delete *i* from $C_3$. Now, we have the final search list $F_3 = \{a, b, c, j, k\}$. In Figure (d), we examine each 3-combination of nodes in $F_3$, and find out a 3-node blackhole $\{a, b, c\}$. Therefore, there are two 3-node blackhole patterns in this example, $\{a, b, c\}$ and $\{i, j, k\}$ respectively.

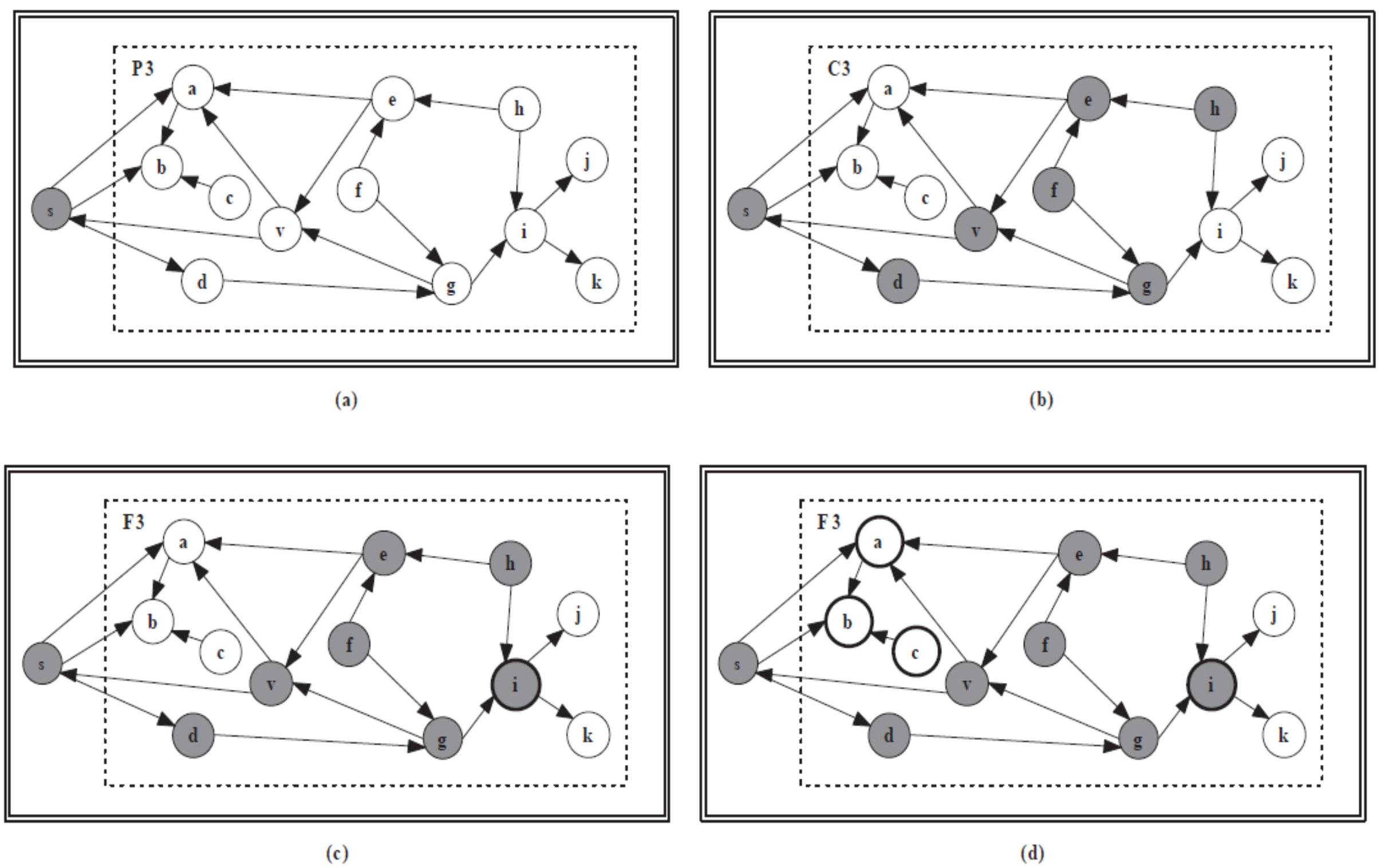

**Figure 8: An example of the procedure of the iBlackhole Algorithm**

## 4.5 The iBlackhole-DC Algorithm

While the search space has been reduced dramatically in the iBlackhole algorithm, it is still possible to develop some pruning strategies for the graphs with some special characteristics. Indeed, for a node $v \in V$, *v* can only form a blackhole pattern with nodes within the same weakly connected component in *G* by the blackhole definition. Therefore, if a directed graph has several weakly connected components, which are not connected to each other, a divide-and-conquer pruning strategy can be exploited for first identifying these weakly connected components and the blackhole finding method can be conducted in each weakly connected component. This pruning strategy can drastically divide a large exponential growth search space into several much smaller exponential growth search space, and thus reducing a lot of computational cost.

In this paper, we combine the iBlackhole algorithm with this divide-and-conquer pruning strategy and develop an even more effective algorithm, named iBlackhole-DC, for finding blackhole patterns. Figure 9 shows the scheme of this algorithm for finding out 1 through *n*-node blackhole patterns in a directed graph *G*.

The **completeness and correctness** of the iBlackhole-DC algorithm is straightforward. Since the only difference between iBlackhole and iBlackhole-DC is the use of the divide-and-conquer strategy. We know that the iBlackhole algorithm is complete and correct. Also, the divide-and-conquer strategy only separates the nodes which cannot form blackhole patterns. Therefore, the iBlackhole-DC algorithm is also complete and correct.

```
ALGORITHM iBlackhole-DC(G = (V, E), n)
Input:
        G: the directed graph
        V: the set of all nodes
        E: the set of all edges
        n: max number of nodes each blackhole may contain
Output:
        Blackhole: 1 to n-node blackhole set of G

        Blackhole ← ∅
        for i ← 1 to n do
            establish potential list P_i
            remove useless nodes from P_i, get candidate list C_i
            remove useless nodes from C_i, get final list F_i
            for each weakly connected component in G(F_i) do
                apply the Brute Force Algorithm to
                   find out the i-item blackhole set {B}
                Blackhole ← Blackhole ⋃ {B}
            end for
        end for
        return Blackhole
```

**Figure 9: A scheme of the iBlackhole-DC algorithm**

## 5. EXPERIMENTAL RESULTS

Here, we present the experimental results to evaluate the performances of iBlackhole and iBlackhole-DC algorithms.

### 5.1 The Experimental Setup

**Experimental Data.** The experiments were conducted on four real-world data sets: *Wiki*, *Amazon*, *Roget*, and *Stock*. Table 1 shows some characteristics of these data sets.

Table 1: Data characteristics

| Data set | # nodes | # egdes |
|---|---|---|
| Wiki | 7,115 | 103,689 |
| Wiki500 | 500 | 3,865 |
| Wiki1000 | 1,000 | 9,741 |
| Wiki1500 | 1,500 | 16,389 |
| Wiki1500-full | 1,500 | 16,820 |
| Amazon | 262,111 | 1,234,877 |
| Amazon1000 | 1,000 | 3,952 |
| Amazon500-full | 500 | 1,911 |
| Roget | 1,022 | 5,075 |
| Roget-full | 1,022 | 5,127 |
| Stock-0.35 | 2,453 | 273 |

**The *Wiki* Data Set.** There are 7,115 nodes and 103,689 edges in the *Wiki* data set [14]. To make the brute-force algorithm runnable, we derived three subgraphs from the original graph, with the number of nodes 500, 1,000, and 1,500 respectively. These subgraphs are named as *Wiki500*, *Wiki1000*, and *Wiki1500* separately. In addition, we synthesized a weakly connected directed graph, named as *Wiki1500-full*, by adding some edges to *Wiki1500* data set.

**The *Amazon* Data Set.** There are 262,111 nodes and 1,234,877 edges in the *Amazon* data set [13]. Similar to the *Wiki* data set, we derived a subgraph *Amazon1000* with 1000 nodes, and synthesized a weakly connected directed graph *Amazon500-full* from the original graph.

**The *Roget* Data Set.** There are 1,022 nodes and 5,075 edges in the *Roget* data set [3]. Also, we synthesized a weakly connected directed graph *Roget-full* from the original graph, by adding some edges to it.

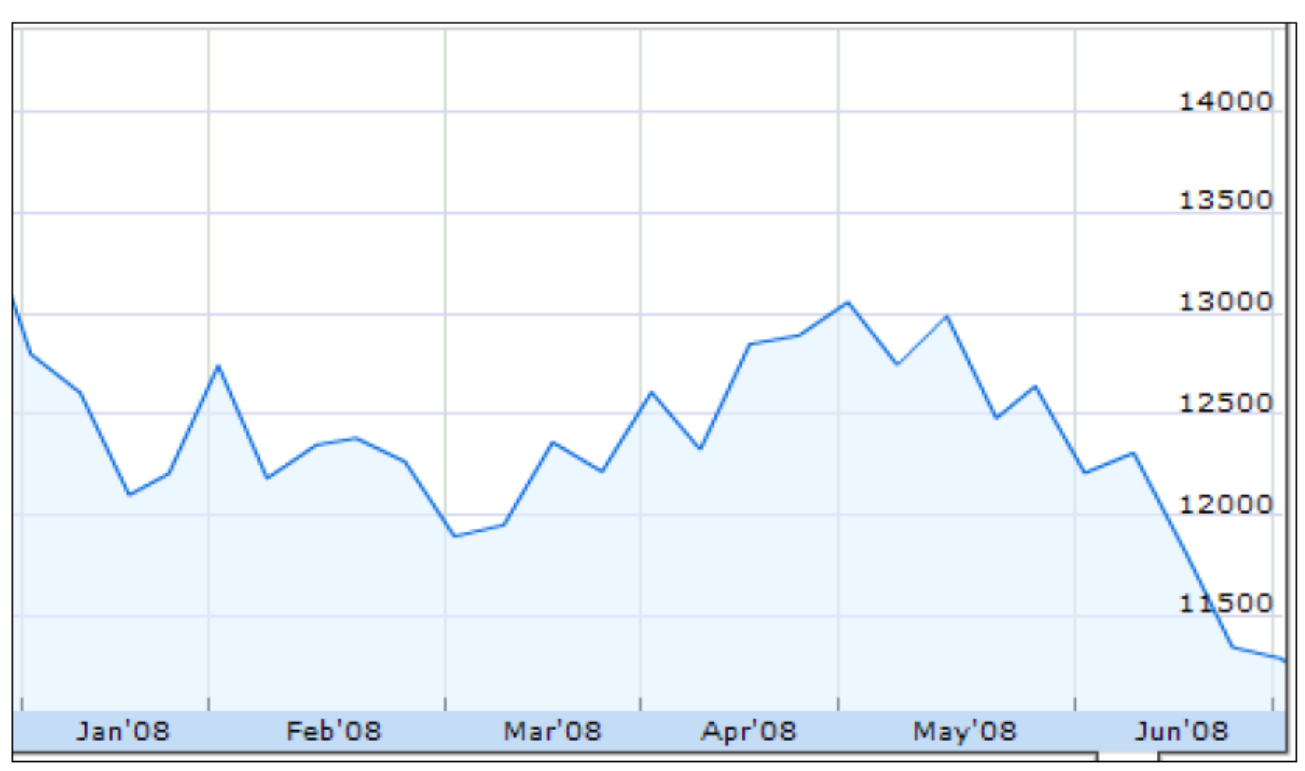

**Figure 10: An overview of the Dow Jones Index from January, 2008 to June, 2008**

**The *Stock* Data Set.** We also generated a *Stock* network data set. Specifically, we collected daily stock prices from Wharton Research Data Services [1] of 3,081 instruments in the U.S. stock market over a period of 125 consecutive trading days from Jan 2, 2008 to Jun 30, 2008. We tried to avoid selecting the period with a strong movement trend in the stock market, since the movements of all instruments during that period tend to have high correlations among each other. As can be seen in Figure 10, there was no strong trend in the Dow Jones index during the selected period. Then we removed instruments in Dow Jones and S&P 500 indexes from our collection. Those instruments are more representative in the stock market and therefore tend to have high correlations with the other instruments. Since we target on finding out some not-so-obvious blackhole patterns, we only consider instruments not in Dow Jones and S&P 500 indexes. After that, we constructed the *Stock* data set as follows: **1)** Nodes in this data set correspond to instruments. There are 2,453 nodes in this data set; **2)** we build a vector $P_i = \{p_{i1}, p_{i2}, \ldots, p_{it}, \ldots, p_{i125}\}$ for each instrument, where $p_{it}$ is the closing price of instrument $i$ on day $t$; **3)** we create a Boolean vector $B_i = \{b_{i1}, b_{i2}, \ldots, b_{it}, \ldots, b_{i124}\}$ based on $P_i$, where $b_{it} = 1$, if $p_{it+1} \geqslant p_{it}$, otherwise 0; **4)** For $X = \{x_1, \ldots, x_t, \ldots, x_n\}$ and $Y = \{y_1, \ldots, y_t, \ldots, y_n\}$, $\rho_{xy}(k)$ is the lagged correlation when $Y$ is delayed by $k$. A symmetric situation can be applied to get $\rho_{yx}(k)$. We compute the lagged correlations $\rho_{ij}(1)$ and $\rho_{ji}(1)$ for each pair of instrument $i$ and $j$; **5)** there is an edge from node $j$ to node $i$, if $\rho_{ij}(1) > \theta$, where $\theta$ is a pre-specified threshold. Since we compute the lagged correlation of 1-day delay between two instruments, if there is an edge from node $j$ to node $i$, it indicates that the movement of instrument $j$ followed the movement of instrument $i$ on the previous day. Here, we specify $\theta$ as 0.35 to get the *Stock-0.35* data set.

Note that the method we used to construct the *Stock* data set is similar to the way in [4]. However, there are some differences. We used the lagged Pearson correlation among instruments, and ended up with a directed graph. While Boginski et al [4] employed the general Pearson correlation and constructed an undirected graph.

**Experimental Platform.** All the experiments were performed on a Dell Optiplex 960 Desktop with Intel Core 2 Quad Processor Q9550 and 4 GB of memory running the Windows XP Professional Service Pack 3 operating system.

### 5.2 An Overall Comparison

In this subsection, we provide an overall comparison of Brute-Force, iBlackhole, and iBlackhole-DC algorithms.

First, we compare the performances of three algorithms on different data sets with almost the same number of nodes. In this experiment, we choose data sets *Wiki1000*, *Amazon1000*, and *Roget*. Figure 11 shows the running time of these algorithms. As can be seen, both Brute-Force and iBlackhole algorithms are runnable within certain number of nodes, while iBlackhole can go a litter bit further than Brute-Force. In contrast, the iBlackhole-DC algorithm is runnable for finding $n$-node blackhole patterns with a large $n$ value.

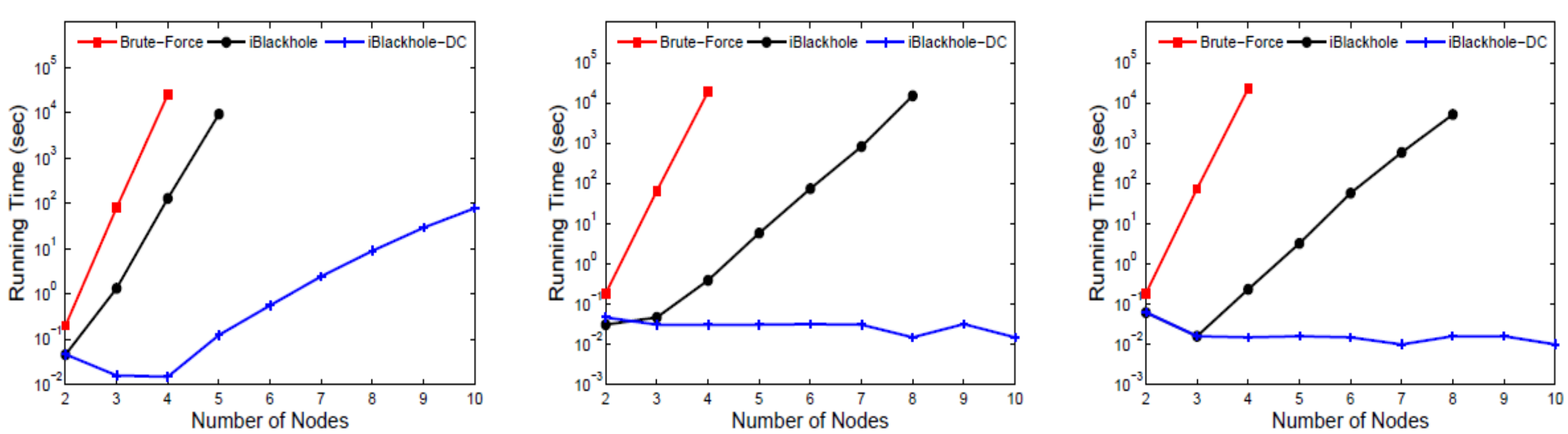

(a) Wiki1000 (b) Amazon1000 (c) Roget

**Figure 11: Running time of Brute-Force, iBlackhole, and iBlackhole-DC on different data sets**

The running time of three algorithms for detecting blackhole patterns with different number of nodes forms three approximately straight lines in logarithm scale for all three data sets. (For iBlackhole-DC, it is more clear if we only focus on $n \geqslant 4$). This indicates that the running time for those algorithms follow an exponential increasing time. Also, the slopes of three performance curves for each data set are significantly different. For Brute-Force, since we do the exhaust search at the beginning and the number of nodes of the three data sets are almost the same, the slopes in those three subfigures are almost the same. For iBlackhole, as well as iBlackhole-DC, they are a little different. The slope of the curve on the *Wiki1000* data set is larger than slopes in *Amazon1000* and *Roget*. For both iBlackhole and iBlackhole-DC, we prune irrelevant nodes from each data set. However, the pruning effect depends on the graph properties of each data set (i.e. the average in-degree and out-degree plays an important role). This makes the running time of iBlackhole and iBlackhole-DC algorithms vary for different data sets, but after all, much less than the Brute-Force algorithm.

Next, we compare the performances of three algorithms on the same data set with different number of nodes. In this experiment, we choose data sets *Wiki500*, *Wiki1000*, and *Wiki1500*. Figure 12 shows the running time of these three algorithms on those three data sets.

The overall performances of these three algorithms are very similar to the first experiment. However, there are still something interesting here. We can observe that the slopes of the three lines in these three data sets are almost the same. (For iBlackhole-DC, it is more clear if we only focus on $n \geqslant 4$). Since these three subgraphs are derived from the same network, the inherent graph properties of these data sets are similar. The above might be the reason that similar slopes are observed in the results.

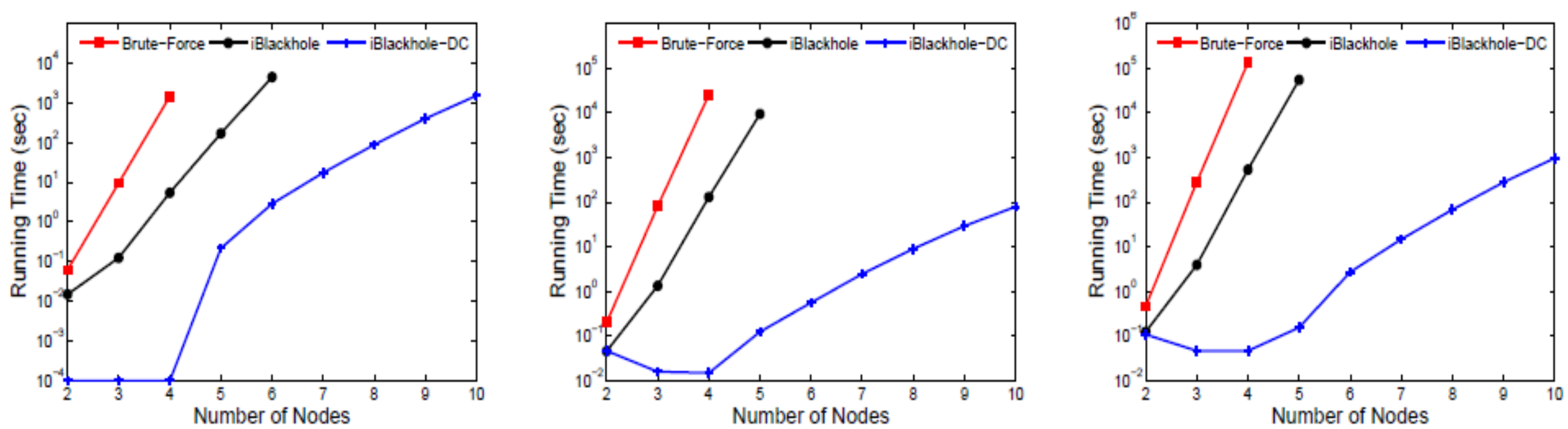

(a) Wiki500 (b) Wiki1000 (c) Wiki1500

**Figure 12: Running time of Brute-Force, iBlackhole, and iBlackhole-DC for different # nodes**

### 5.3 iBlackhole vs. iBlackhole-DC

In this subsection, we compare the performances of iBlackhole and iBlackhole-DC algorithms. We show how significant the divide-and-conquer strategy improves the performance of iBlackhole. In this experiment, we choose three synthetically weakly connected directed networks, *Amazon500-full*, *Roget-full*, and *Wiki1500-full*.

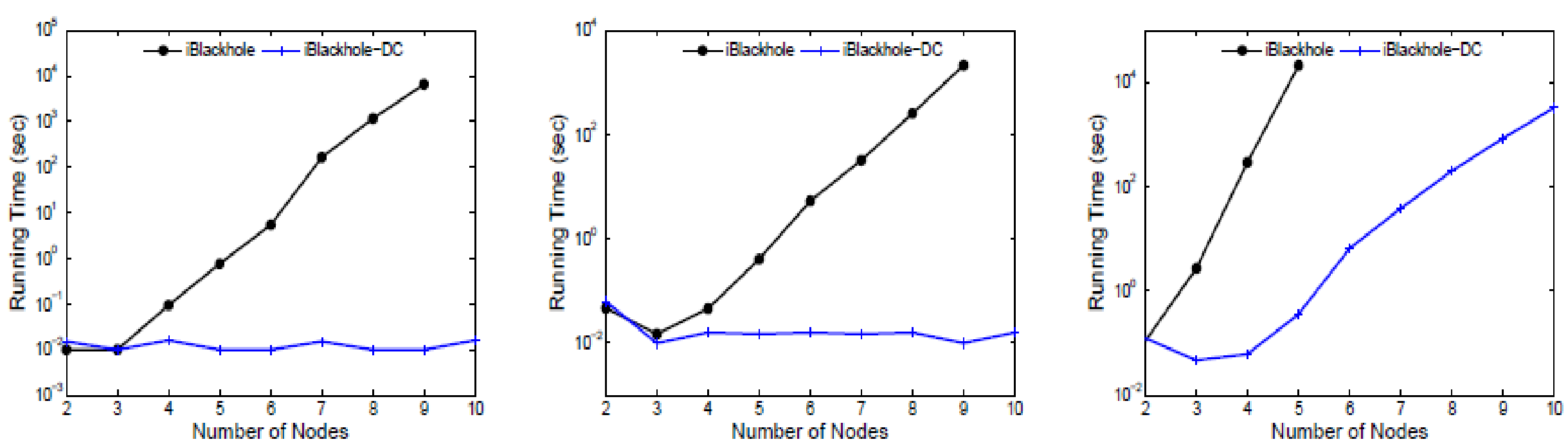

(a) Amazon500-full (b) Roget-full (c) Wiki1500-full

**Figure 13: The running time of iBlackhole and iBlackhole-DC algorithms on different data sets**

Figure 13 shows the running time of these two algorithms on those three data sets. In the figure, we can see that the performance of iBlackhole-DC is several orders of magnitude faster than the performance of iBlackhole, since it drastically divides a large exponential growth search space into several much smaller exponential growth search space, and thus reduces a lot of computational cost.

Figure 14 shows the visualizations of the structures of different data sets before and after applying the first pruning scheme to these data sets while detecting 7-node blackhole patterns. This figure is drawn with Pajek [17]. From this figure, we can observe that the number of nodes in each data set decreases dramatically after pruning, and each network becomes very sparse. Table 2 shows some main characteristics of the data sets after pruning. As can be seen, while the original data sets are all weakly connected, we can still get a large number of connected components after pruning. Therefore, the divide-and-conquer strategy can help dramatically reduce the search space.

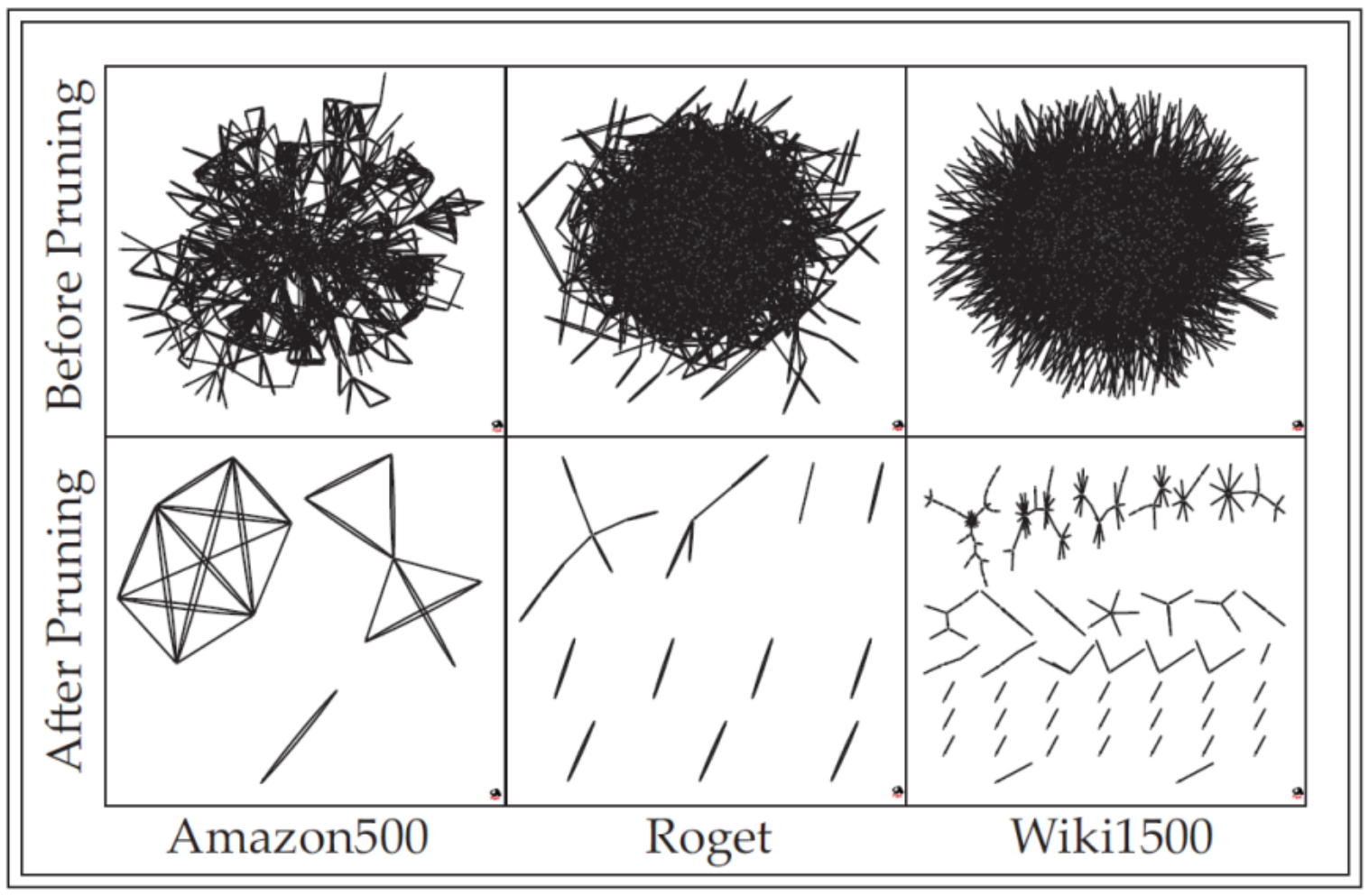

**Figure 14: Visualizations of structures of different data sets before and after pruning**

Table 2: Characteristics of data sets after pruning

| Data set | # nodes | # edges | # connected comp |
|---|---|---|---|
| Amazon500 | 14 | 36 | 3 |
| Roget | 32 | 37 | 11 |
| Wiki1500 | 252 | 209 | 43 |

## 5.4 Blackhole Patterns in the Stock Data

Here, we show an application of blackhole patterns for understanding the structural relationship of stock movement.

Figure 15 shows two blackhole patterns identified in the *Stock-0.35* data set. Owens Corning (ticker: OC) is in the left blackhole pattern. The Westmoreland Coal Company (ticker: WLB) has an outlink to OC. This indicates that the price movement of WLB followed the price movement of OC. By doing some research, we find out Owens Corning is one of the biggest building material producers in the country. Its products include the manufactured stone products used in the building. In recent years, there is a trend in the industry that companies are developing new innovative building materials by recycling the waste in the energy industry, which are primarily the residual byproducts of coal combustion. As an energy company, WLB owns five coal mines. Therefore, it is understandable that the stock price of the Westmoreland Coal Company has a lag correlation with the stock price of Owens Corning. The other two companies in this pattern are Venoco Inc.(ticker: VQ) and Helmerich & Payne Inc. (ticker: HP). Venoco Inc. [2] is an energy company primarily engaged in the acquisition, exploration, exploitation and development of oil and natural gas properties, while Helmerich & Payne Inc. [2] is a contract drilling company drilling oil and gas wells for others. Therefore, it is not surprising that the stock price movements of these two companies are lag correlated with the stock price of the Westmoreland Coal Company. Indeed, the blackhole patterns can help illustrate this type of structural relationships of stock movements of several companies.

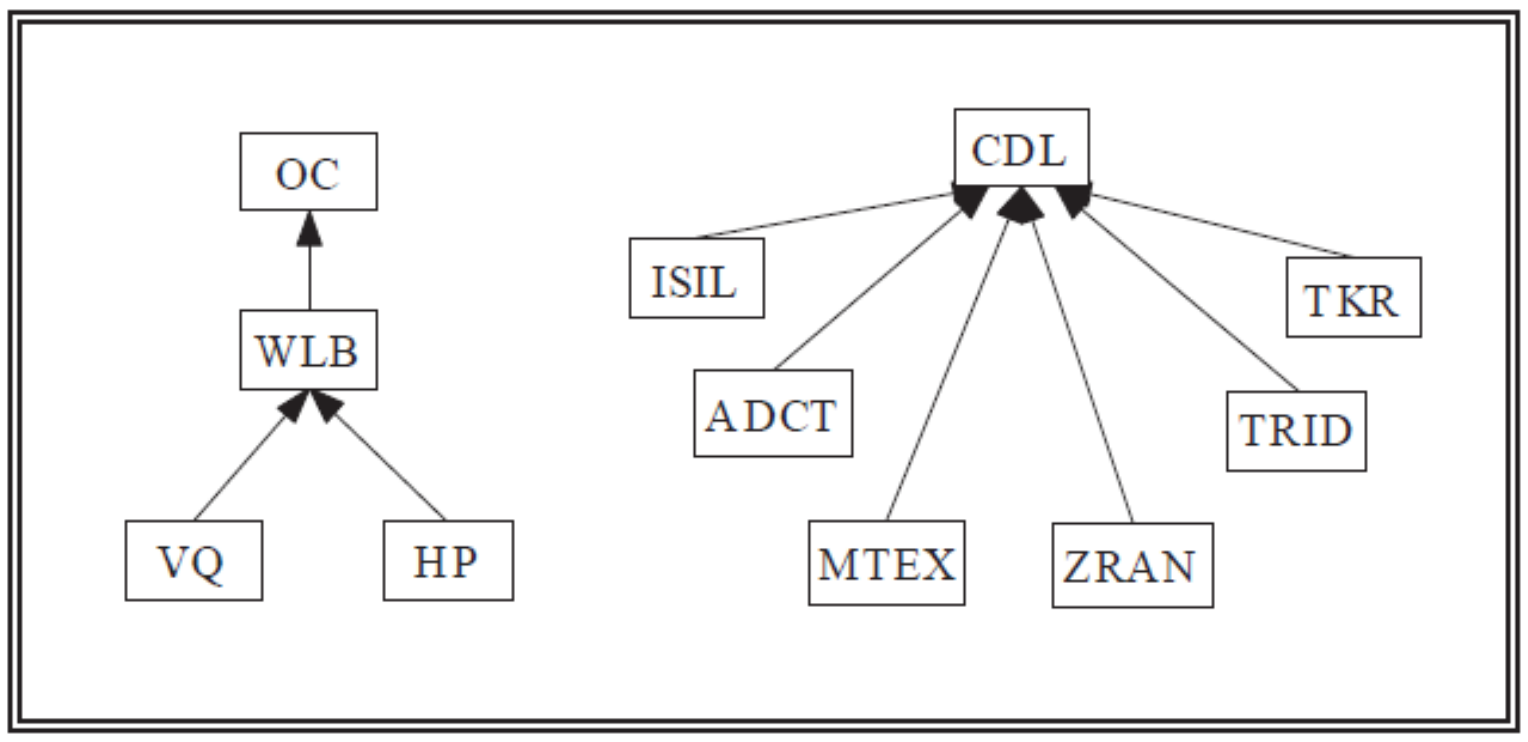

**Figure 15: Illustration: two blackhole patterns identified in the *Stock-0.35* data set**

The second blackhole pattern is a star-shaped blackhole, which indicates the stock prices for the other six instruments are triggered by Citadel Broadcasting Corp. (ticker: CDL). Among these six companies, there are one telecommunication company (ADCT), three IC related companies (ISIL, TRID, and ZRAN), and one highly engineered steel produce company (TKR), which are all related to the Broadcasting Corporation to some extent. The other company is a wellness solution provider, which may be involved in this pattern by chance or for some unknown reasons.

This application is just a simple indication of the use of the blackhole patterns. Indeed, the blackhole patterns can provide an unique view of some structural properties, and help us better understand the interactions among some nodes in the network. However, we should note that this use of blackhole patterns is still preliminary and more comprehensive studies are expected in the future.

## 6. RELATEDWORK

Related work can be grouped into two categories. The first category includes the work on frequent subgraph mining, which studies how to efficiently find frequent subgraphs in the graph data. For instance, Jiang et al. [11] proposed a measure for mining globally distributed frequent subgraphs in a single labeled graph. Meanwhile, there are many works in mining frequent subgraphs in multiple labeled graphs [20, 18, 10, 19, 12, 6]. The problem of detecting blackhole patterns is different from the above works for two reasons. First, the definition of blackhole patterns is different from the definition of frequent subgraphs. Second, blackhole patterns are identified whether they are frequent or not.

The second category includes the works for detecting community structures in large networks. Communities in a network are groups of nodes within which connections are dense, but between which connections are sparse [15]. There are a lot of works on how to detect communities in a network. For instance, Newman and Girvan [16, 8] proposed a betweenness-based method, Hopcroft [9] proposed a stable method, and Ghosh [7] proposed a global influence based method to detect community structures. All these methods detect community structures based on certain definitions and criteria. However, the definition of blackhole patterns is different from the above definitions of communities. Also, once a network has been decided, the number of $n$-node blackhole patterns is determined. In contrast, it is usually difficult to know how many community structures are in the network.

## 7. CONCLUDING REMARKS

In this paper, we formulated a problem of finding blackhole and volcano patterns in directed networks. Both blackhole and volcano patterns can be observed in real-world scenarios, such as the trading ring for market manipulation. Indeed, it is essentially a combinatorial problem for mining blackhole or volcano patterns. To reduce the complexity of the problem, we first proved that the problem of finding blackhole patterns is a dual problem of finding volcano patterns. Thus, we could be only focused on mining blackhole patterns. To that end, we derived two pruning schemes. The first scheme is based on a set of size-independent pruning rules which can help to prune the candidate search space effectively and thus can dramatically reduce the computational cost of blackhole mining. Based on the first pruning scheme, we developed the iBlackhole algorithm for mining blackhole patterns. In addition, the second scheme is to take advantage of an unique graph property; that is, we could search in each individual subgraphs if the target directed graph contains several disconnected subgraphs. Therefore, by exploiting these two pruning schemes, we developed the iBlackhole-DC algorithm for finding blackhole patterns.

Finally, as shown in the experimental results, the pruning effect of both pruning schemes is significant and the iBlackhole-DC algorithm is several order-of-magnitude faster than the iBlackhole algorithm, which outperforms a bruteforce approach by several orders of magnitude as well.